\def\year{2020}\relax
\documentclass[letterpaper]{article} 
\usepackage{aaai20}  
\usepackage{times}  
\usepackage{helvet} 
\usepackage{courier}  
\usepackage[hyphens]{url}  
\usepackage{graphicx} 
\usepackage{graphicx}  

\usepackage{amsfonts}
\usepackage{amsmath}
\usepackage{algorithm, algorithmic}
\usepackage{booktabs}
\usepackage{multirow}

\title{Self-Paced Video Data Augmentation with Dynamic Images Generated by Generative Adversarial Networks}
\author{Yumeng Zhang\textsuperscript{\rm 1}, Gaoguo Jia\textsuperscript{\rm 1}, Li Chen\textsuperscript{\rm 1}, Mingrui Zhang\textsuperscript{\rm 2}, Junhai Yong\textsuperscript{\rm 1}\\
\textsuperscript{\rm 1}Tsinghua University, Beijing, China\\
\textsuperscript{\rm 2}Beijing University of Posts and Telecommunications, Beijing, China
} 
\begin{document}

\maketitle

\begin{abstract}

There is an urgent need for an effective video classification method by means of a small number of samples. The deficiency of samples could be effectively alleviated by generating samples through Generative Adversarial Networks (GAN), but the generation of videos on a typical category remains to be underexplored since the complex actions and the changeable viewpoints are difficult to simulate. In this paper, we propose a generative data augmentation method for temporal stream of the Temporal Segment Networks with the dynamic image. The dynamic image compresses the motion information of video into a still image, removing the interference factors such as the background. Thus it is easier to generate images with categorical motion information using GAN. We use the generated dynamic images to enhance the features, with regularization achieved as well, thereby to achieve the effect of video augmentation. In order to deal with the uneven quality of generated images, we propose a Self-Paced Selection (SPS) method, which automatically selects the high-quality generated samples to be added to the network training. Our method is verified on two benchmark datasets, HMDB51 and UCF101. The experimental results show that the method can improve the accuracy of video classification under the circumstance of sample insufficiency and sample imbalance.

\end{abstract}

\section{Introduction}
Video classification has broad application prospects in security, social media and many other fields. At present, the researches on video classification mainly explore how to effectively utilize video inter-frame information \cite{wang2016temporal}, which use powerful deep learning method and rely on sufficient training data. In some special application scenarios, such as illegal video monitoring or minority video classification, the number of samples in some categories is very small, and these methods are difficult to achieve satisfactory classification results.

How to train the network with insufficient samples effectively has always been a big challenge for deep learning. At present, the method of processing few data for the video classification field mainly relys on learning few-shot video representation using the memory network \cite{zhu2018compound} and long-term temporal ordering information through the Temporal Alignment Module \cite{cao2019few}, so as to improve the utilization efficiency of the sample, and enhance the accuracy of video classification in the case of few samples. However, due to the limited quantity of samples and the nature of neural network, overfitting is somewhat inevitable. One possible solution for the overfitting problem is to do data augmentation taking advantage of powerful generating capabilities of Generative Adversarial Networks \cite{goodfellow2014generative}, which has been extensively studied in the field of Person Re-Identification \cite{zhong2018camera,liu2018pose,zheng2017unlabeled} and medicine image processing \cite{gao2019data,frid2018synthetic}. The data generated by GAN can serve as a regularizer to alleviate the over-fitting problem as well as enhance the features.

\begin{figure}[t]
    \centering
    \includegraphics[width=0.45\textwidth]{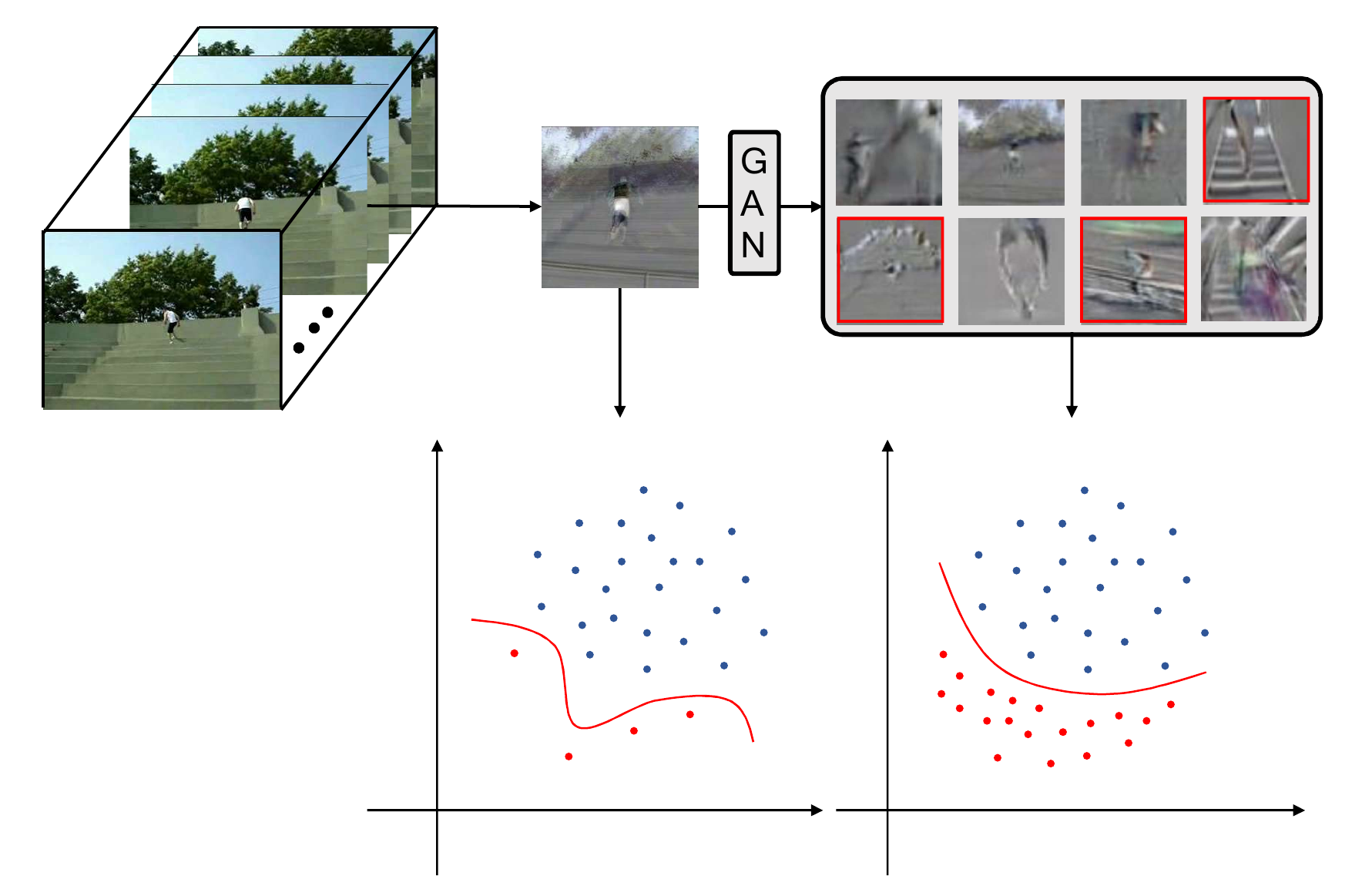}
    \caption{The abridged general view of our method. When there are insufficient samples to train the network, it is difficult to get a satisfactory decision surface. Our method alleviates the problem by generating multiple dynamic images with GAN. We select several high quality generated samples (e.g. the images in red boxes) to modify the decision surface, which improves the accuracy of video classification.}
    \label{fig:breifintroduction}

\end{figure}

However, since videos are more complicated than images, the generation of videos on a typical category remains to be an under-researched task. Acharya et al. \cite{acharya2018towards} introduced a progressive growing GAN, which generated videos of 256x256x32 resolution for the first time. Saito et al. \cite{saito2018tganv2} proposed Temporal GAN v2 and achieved a significantly improvement of Inception Score on the UCF101 datasets over the previous methods. However, it is difficult to generate the videos of a typical category with complex temporal actions and changeable sences using the methods as these videos contain huge information which requires extremely high memory capacity. The current networks can hardly meet the needs. Thus it is impracticable to use these methods as a data augmentation for video classification. In general, the method of generative augmentation for video data has not been profoundly researched, and this is what and why we focus and explore.
 
\begin{figure*}[ht]
    \centering
    \includegraphics[width=1\textwidth]{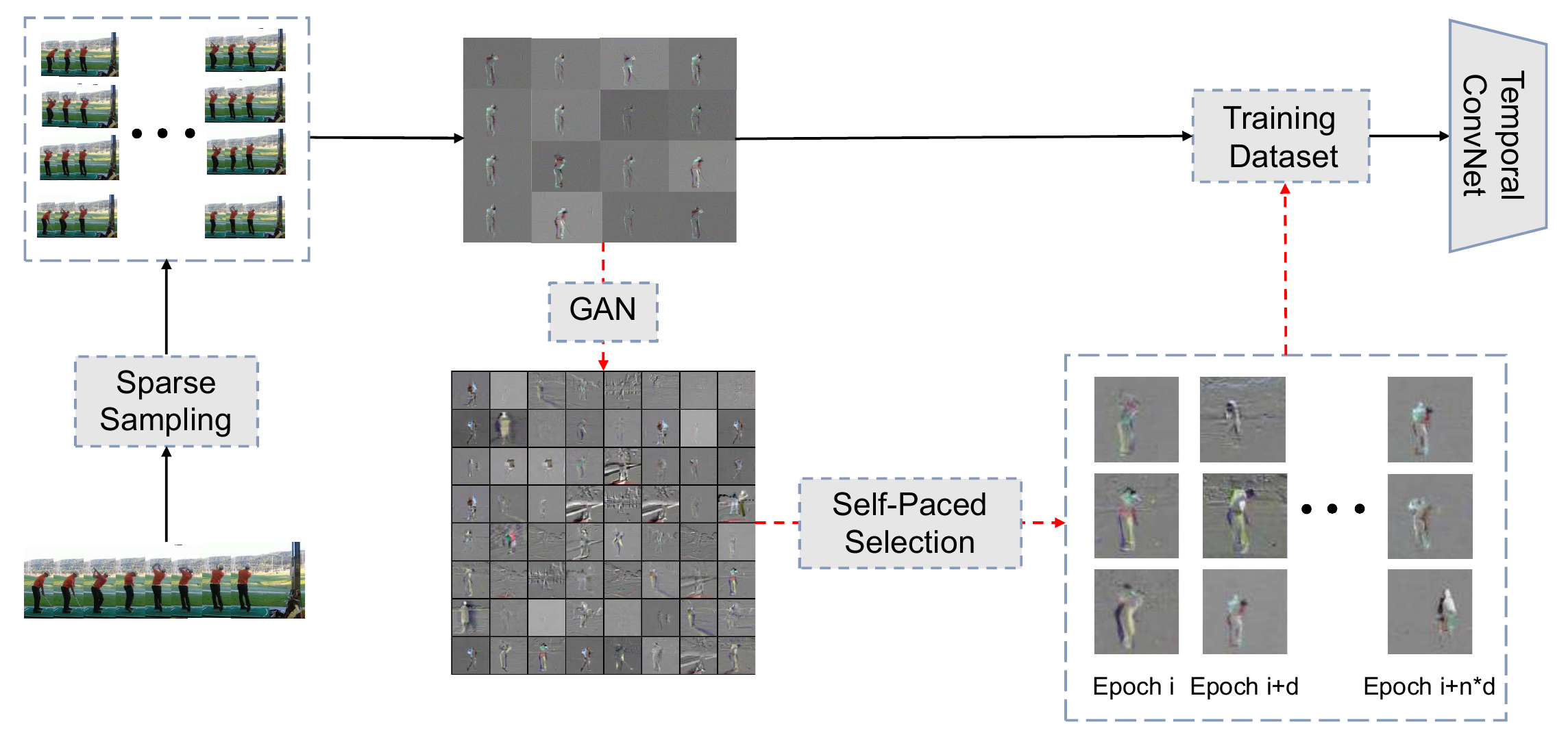}
    \caption{The overview of our Method. We first break the videos into several segments by sparse sampling to obtain enough dynamic images for training of GAN. The method selects a fixed number of generated samples by SPS at different epochs $\left\{ i,\cdots,i+n \times d \right\}$, adding them to the training dataset for Temporal ConvNet, where the i represents the starting epoch, d is the interval and $n+1$ is the total number of epochs using SPS. }
    \label{fig:wholenetwork}
\end{figure*}

We stand on a more novel perspective, rather than design an algorithm to synthesis videos to solve the problem. Since it is difficult to perform data augmentation for spatial stream of Two-Stream method, which is the most commonly used method in the field of video classification, we focus on generating samples for temporal stream based on the idea of dynamic image \cite{bilen2016dynamic}. The dynamic image compresses the motion information of the video into a still image, which maintains the important information with small memory usage. The interference factors such as background are removed as they tend to be static in the video. Therefore, it is easier for GAN to grab the main characteristics of the videos. The training of GAN requires a large quantity of samples, while there are only a small number of the videos in the case of sample insufficiency. In order to solve the problem, we adopt a sparse sampling technique to get sufficient dynamic images with subvideos. Then, we could train the GAN and synthesize a large number of dynamic images by GAN. 

As is known, the quality of samples generated by GAN is uneven. Several methods adopt label smooth regularization (LSR) \cite{zhong2018camera} to solve the problem by assigning small values to the non-ground truth classes instead of 0, but ignore the fact that poor quality of GAN samples may exert negative impact on training. In order to pick out the generated samples that are most helpful for video classification, we propose a sample filtering strategy called Self-Paced Selection (SPS). We find that the dynamic image contains the clear category information of the video without interference factors, thus the generated samples of poor quality can be easily distinguished by identifying the information. The module selects the generated samples with clear category information at different time steps, adding them to the training set to reduce interference from noise samples. As shown in Figure \ref{fig:breifintroduction}, the generated dynamic images can be used to smooth the decision surface and regularize the network, which significantly improve the accuracy of temporal stream as well as the accuracy of two stream video classification, thereby to achieve the effect of video augmentation.

In a word, the main contributions of this paper are as follows:

\begin{itemize}
    \item We propose a generative data augmentation framework for video classification using the dynamic image. It is the first attempt-to the best of our knowledge-to adopt GAN for video data augmentation.
    \item We propose Self-Paced Selection (SPS) to reduce the impact of noisy samples exert on training.
    \item We augment the samples of temporal stream to enhance the accuracy of video classification. The experiments are conducted on two benchmark datasets UCF101 and HMDB51. The experimental results show that our method significantly improves the performance under the circumstance of sample insufficiency and sample imbalance. 
\end{itemize}

\section{Related Works}
\subsection{GAN Application in Data Augmentation}
The generative adversarial network \cite{goodfellow2014generative} is an important generative model. Through the cooperation of the generator and the discriminator, the GAN can generate high quality samples from random noise with the training data. Models like  DCGAN \cite{radford2015unsupervised}, WGAN \cite{arjovsky2017wasserstein} and SNGAN \cite{miyato2018spectral} were derived from GAN of Goodfellow et al. \cite{goodfellow2014generative}, which significantly improved the quality of generated images. With the rapid development of GAN, some researchers have begun to study the application of GAN in data augmentation. Zheng et al. \cite{zheng2017unlabeled} used DCGAN to generate some unlabeled samples and regularize the network by giving the unlabled samples uniform annotations. Zhong et al. \cite{zhong2018camera} proposed CamStyle GAN to smooth the camera style disparities. Huang et al. \cite{huang2018auggan} introduced AutoGAN for scene transformation, to realize data enhancement. Although GAN has made certain achievements in data augmentation of image, video generation remains to be an under-researched problem. As the video data contains specific timing information, the GAN for video generation can only generate simple presentation photography or small scenes \cite{bansal2018recycle}, which is difficult to expand the video dataset in practical applications. So far, there has been no technology that adopts GAN for video data augmentation. Our method is to fill this gap.

\subsection{Video Classification}
Thanks to powerful deep learning technology, video classification technology has achieved unprecedented development. The two stream networks are the most commonly used structure for video classification. Simonyan et al. \cite{simonyan2014two} introduced the optical flow into the video classification. TSN \cite{wang2016temporal} and I3D \cite{carreira2017quo} used 2D CNN and 3D CNN to calculate RGB and optical flow information, which effectively improves the accuracy of video classification. Peng et al. \cite{peng2018two} proposed collaborative learning to capture the static and motion information simultaneously. Although optical flow can be used to represent the motion information effectively, calculating optical flow takes a lot of time.

Some researchers have studied the accelerated calculation of optical flow, such as flownet v2 \cite{ilg2017flownet}, which has achieved a good effect of accelerating optical flow calculation. Others have begun to study alternative methods of optical flow. Zhang et al. \cite{zhang2016real} proposed motion vector which extracts motion information in the form of video compression, but the motion vector is quite noisy and precision is unsatisfactory. Zhu et al. \cite{zhu2017deep} put forward the deep feature flow, but the feature flow relies heavily on the selection of key frames, which may ignore important information due to wrong selection. MotionNet \cite{zhu2018hidden} integrated the extraction of motion information into end-to-end training of the network, while the hidden representation makes it difficult to perform generative data augmentation as they only use videos as input. Bilen et al. \cite{bilen2016dynamic} introduced the idea of dynamic image, compressing the motion information of the whole video into a still image. The dynamic image generalizes well to the 2D convolution network, which motivates us to develop our algorithm based on it. The goal of Bilen's method is to accelerate the algorithm with dynamic images, while in this paper, we alleviate the problem of insufficient samples with generated dynamic images by GAN. The dynamic image shows the motion information of the video, thus the generated image containing the similar motion information corresponds to a video of certain category. The generated dynamic images indirectly expand the number of video samples, which can be used to smooth the decision surface and regularize the network.

\section{Methods}

\subsection{OverView}

We aim to use generated dynamic images of high quality to achieve the effect of video augmentation. The diagram of our algorithm is shown in Figure \ref{fig:wholenetwork}. Since it can hardly meet the requirements of the training of GAN under the circumstance of sample insufficiency, we first get multiple dynamic images of subvideos obtained by sparse sampling technique. In order to avoid the gradient vanishing and mode collapse problems, we use the WGAN to generate samples of specific video type. It is known the quality of generated samples by GAN varies in a large range and the generated samples of poor quality may exert negative impacts on training, thus we propose a Self-Paced Selection method to automatically filter the samples. The module estimates the quality of all generated samples with a well-designed selection indicator at specific epochs, and a fixed number of generated samples that are most helpful for training are added to the training set. In the following sections, we will introduce our method in detail. 

\subsection{Generative Adversarial Networks for Dynamic Image}

\textbf{Dynamic image} Dynamic image aims to represent a video $I= \left\{I_1,\cdots,I_T \right\}$ as a still image $d^* \in \mathbb{R}^{m \times n}$, where T is the total number of frames and (m,n) represents the resolution of a single frame. It is proposed by \cite{bilen2016dynamic}. We now introduce this algorithm briefly. Given a component-wise mapping function $\psi(\cdot)$, a score S(t$|$d) is defined for frames $\left\{I_1, \cdots, I_t \right\}$ to reflect the rank of the frames in the video, where later times have a higher score. The definition of score is shown in Eq. \ref{eq:score}. 
\begin{equation}
S(t|d) = <d,V_t>
\label{eq:score}
\end{equation}
where $V_t = \frac{1}{t} \sum \psi(I_i)$. The dynamic image $d^*$ is learned from the following convex optimization problem.

\begin{equation}
    \begin{aligned}
    &E(d) = \lambda\left \| d^2 \right\| + \\
    & \qquad \frac{2}{T(T-1)} \times \sum_{q>t} max(0, 1-S(q|d)+S(t|d)) \\
    &d^* = argmin_{d} \; E(d)
    \end{aligned}
\end{equation}

\textbf{Sparse sampling} Since the training of GAN model requires a large number of samples, we adopt a sparse sampling strategy to obtain sufficient dynamic images. We randomly select half number of frames in a video to form a new video, which is used to calculate the simplified dynamic image. We repeat this process many times to get large amounts of dynamic images. 

\textbf{Wasserstein generative adversarial networks} A generative adversarial network (GAN) is a class of machine learning systems. The generator and discriminator networks contest with each other in a game to generate new data with the same statistics as the training set. It is well known that the conventional GAN is difficult to train due to the vanishing gradient problem. Wasserstein generative adversarial networks utilize the wasserstein distance to cure this training problem, which simplifies the training difficulty and reduces the mode collapse. Thus we adopt this method to generate new dynamic images. The definition of wasserstein distance is shown in Eq. \ref{eq:wassersteindistance}.
 
\begin{center}
    \begin{equation}
    \begin{split}
    W(p_r,p_\theta) = &\underset{\zeta\in\gamma}{\inf}  \underset{x,y}{\iint}\|x-y\| \zeta(x,y)dxdy\\ 
    &=  \underset{\zeta\in\gamma}{\inf} \mathbb{E}_{x,y~\zeta}[\|x-y\|]
    \end{split}
    \label{eq:wassersteindistance}
    \end{equation}
\end{center}

where the $p_r$, $p_\theta$ are two continuous distributions and $\gamma(p_r, p_\theta)$ are their joined distributions. In our case, the $p_r$ is the distribution of real dynamic images obtained by sparse sampling. The $p_\theta$ is initialized to noise distribution and then is drawed closer with the $p_r$ by minizing their wasserstein distance.

As different type of videos have different motion information, it is useless to mix different categories of dynamic images to train WGAN. Thus we train a wgan model for each type of video that needs to be augmented. 

\subsection{Self-Paced Selection}

Self-paced learning is a learning methodology \cite{kumar2010self}, it adaptively prioritize learning simple, reliable examples and gradually move over to difficult samples. 
Given a training dataset as $D = \left\{(x_i,y_i), \cdots, (x_n,y_n)\right\}$ and the current network $f_{W}$, where $x_i \in X$ are the samples and $y_i \in Y$ are the corresponding labels, we first compute the loss function $f(x_i,y_i;w_t)$ and regularization function $r(w_t)$. Then the parameter update process (objection function) can be denoted as,
\begin{equation}
w_{t+1} = argmin(r(w_t) + \sum_{(x_i,y_i)\in D} f(x_i, y_i; w_t))
\end{equation}
The main idea of self-paced learning is to select with small training errors and high likelihood values from all samples in each iteration and the update the model parameters. Therefore, self-paced learning introduces a binary variable $v_i$ in the objection function, which is used to represent whether each sample is selected. Formally, the objection function becomes the formula \ref{eg:selfpaced}
\begin{equation}
\begin{aligned}
    (w_{t+1}, &v_{t+1}) = argmin(r(w_t) + \\
    &\sum_{(x_i,y_i)\in D} v_i f(x_i, y_i; w_t) - \frac{1}{K}\sum_{(x_i,y_i)\in D} v_t)
\end{aligned}
\label{eg:selfpaced}
\end{equation}
K is a weight that determines the number of samples to be considered. 

With the idea of self-paced learning, we design an automatic selection module for generated images. We call $f_w(\cdot)$ the video classification network and $f(x_i,y_i;w)$ the object function. The training net is split into two parts, sparse sampling dynamic images (SSDI) and generated dynamic images (GDI). The SSDI images is used as the initial training set $T_0$. Then we select $N_g$ samples at a specific epoch $E_g^i \in \left\{E_g^1, \cdots, E_g^t\right\}$, and add them to the training set $T_{E_g^{i-1}}$ as $T_{E_g^i}$. The conventional self-paced learning selects the samples based on the loss value. It works well in selecting samples from single datasets. However, the goal of our selection module is to distinguish which generated samples are helpful for video classification and there is a domain gap between the generated samples and sparse sampling samples. Thus there are cases where the loss is small and the sample contains much noise due to the imperfect classifier on unseen generated samples. In view of the above problems, we propose a new metric to select generated samples for training. 

We put all the generated samples to the video classification network, and get a score matrix $S = \left\{ S_{k,n} \right\}$, k = $\left\{ 1,\cdots,K \right\}$, n = $\left\{ 1,\cdots,N \right\}$. where k in the sample index, K is the total number of generated samples and n is the corresponding label, N is the total number of classes. We denote the largest value of $S_k$ as $Max_k^1$ and the second largest value of $S_k$ as $Max_k^2$. If a generated sample is relatively clean, it should have a clear category attribution. Therefore, the classifier can classify it more definitely. This determinism is reflected in the difference between the maximum and the sub-value, which is represented by $Max^1 - Max^2$. The larger the difference, the more obvious the category information. Based on the above ideas, we calculate the $Max^1 - Max^2$ of all generated samples and the add top $N_g$ samples to the training set. The whole process is shown in Algorithm \ref{alg:1} and the parameter update process becomes the formula \ref{eg:sps}.

\begin{equation}
    \begin{aligned}
        &w_{t+1} = argmin(r(w_t) + \\
        &\sum_{(x_i,y_i)\in SSDI} f(x_i, y_i; w_t) + \sum_{(x_i,y_i)\in GDI} g_i f(x_i, y_i; w_t))
    \end{aligned}
    \label{eg:sps}
    \end{equation}

where $g_i$ is a binary variable, which is obtained by SPS to represent whether the $i_{th}$ sample is used for training.

\begin{algorithm}
	\renewcommand{\algorithmicrequire}{\textbf{Input:}}
    \renewcommand{\algorithmicensure}{\textbf{Output:}}
    \caption{Self-Paced Selection}
	\label{alg:1}
	\begin{algorithmic}[1]
        \REQUIRE $GDI, M,N_g,E_g,T_0,W_0$
        \ENSURE Trained network $f_{W_m}$
        \STATE initialize the $T_1$ with $T_0$

        \FOR{epoch :=1 to M}
        \STATE $P := \emptyset \; T_{epoch} := T_{epoch-1}$
        \IF {epoch $\in E_g$:} 
        \FOR {each $ (x_k,y_k)\in GDI:$}
        \STATE compute $D(x_k,y_k) = Max_k^1 - Max_k^2$
        \STATE $P=P\cup((x_k,y_k, D(x_k,y_k)))$
        \STATE Let $P^{epoch}$ be the $N_g$ topmost tuples in P according to the $D(x_k,y_k)$
        \ENDFOR
        \FOR {each $(x_k,y_k,D(x_k,y_k)) \in P^{epoch}:$}
        \STATE $T_{epoch} = T_{epoch} \cup (x_k,y_k)$
        \ENDFOR
        \ENDIF
        \ENDFOR
    \end{algorithmic} 
\end{algorithm}

\section{Experiments}
\subsection{Datasets}
We evaluate our method on two benchmark datasets of action recognition, namely UCF101 and HMDB51.

UCF101 dataset contains 101 types of actions, a total of 13320 videos, which were collected in YouTube. The UCF101 has a rich diversity of actions, with the presence of large variations in camera motion, object appearance, pose, object scale and so on. 

HMDB51 comprises of 51 human action categories and spans over 6766 videos, which are collected from various sources, mostly from movies.

As the goal of our method is to alleviate the problem of the lack of sufficient samples, we select 4 classes from UCF101 and HMDB51 respectively, each with 20 samples, to simulate the situation of insufficient samples.

We choose brush hair, cartwheel, climb stairs and flic-flac of HMDB51 as a new test datasets, which is recorded as HMDB4. GolfSwing, Shotput, Skiing and YoYo of UCF101 are selected to compose a new dataset, which is denoted as UCF4.

\subsection{Implementation Details} 

\textbf{CNN baseline} We adopt the TSN network \cite{wang2016temporal} as the backbone network of our method. All the images are cropped into 224$\times$224 using corner cropping with scalejittering. We insert a dropout layer before the final convolutional layer and set the dropout rate to 0.4. We use stochastic gradient descent with momentum 0.9. The learning rate is set to $10^{-3}$ and decreases to its $\frac{1}{10}$ every 30 epochs. The whole training process stops at 80th epoch. 

\begin{figure}[ht]

    \centering
    \includegraphics[width=0.45\textwidth]{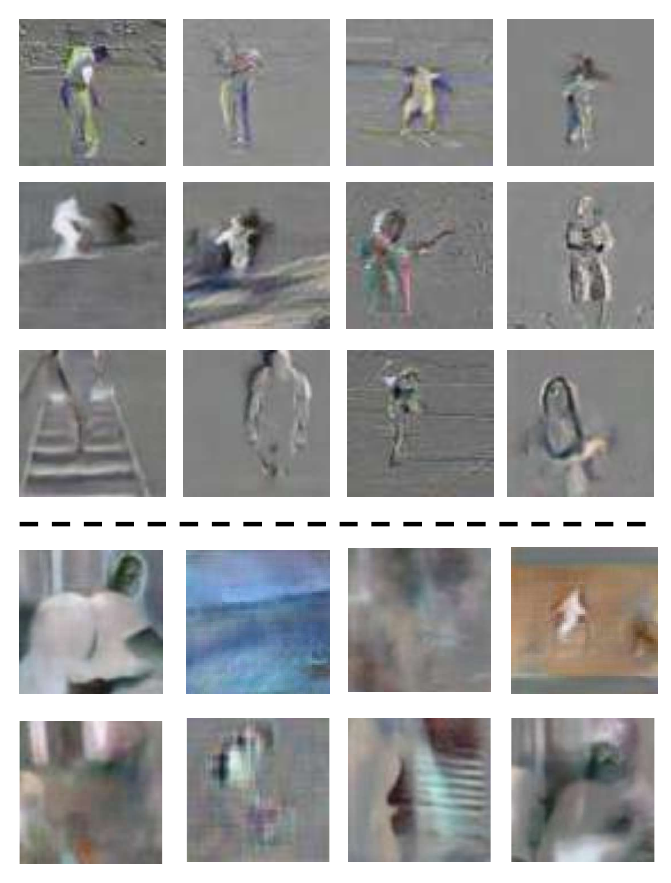}
    \caption{The generated samples of Generative Adversarial Networks. The images above the dotted line are of better visual quality than those below the dotted line. The uneven quality of generated samples motivated to design the SPS, which automaticly distinguishes the samples.}
    \label{fig:generatedsamples}

\end{figure}

\textbf{GAN training and testing} We use Pytorch and WGAN package to train the GAN model using the dynamic image obtained by sparse sampling. The learning rate of adam \cite{kingma2014adam}  is $10^{-4}$. We train the whole network for 20000 iterations. The output images are resized to 256$\times$256 and then used in the training of TSN. The generated images are shown in Figure \ref{fig:generatedsamples}.

\subsection{The Quality of Generated samples}

\textbf{Label Mix Experiment} In order to verify whether the dynamic images generated by the GAN contains information that contributes to the video classification, we first conduct a label mixing experiment with cartwheel and flic-flac videos of HMDB51, since these two type of videos are visually closer. We randomly select $\alpha\%$ samples and give them the wrong labels. Through the experimental results shown in Table \ref{tbl:experiments_mix}, we conclude that the generated samples maintain classification information to a certain extent since $\alpha$ and accuracy are inversely proportional. However, the low accuracy means that there still are lots of noise in generated images, which prompts us to design the SPS.

\begin{table}[htbp]

    \centering
    \setlength{\tabcolsep}{1mm}{
    \begin{tabular}{cccccc} 
    \toprule
    $\alpha$ & 0\% & 50\% & 100\%   \\
      \midrule
      cartwheel and flic-flac &  54.69\% & 51.16\% & 44.75\%  \\

      \bottomrule
      \end{tabular}}
      \caption{The results of Label Mix Experiment. We find that the generated dynamic images contributes to the performance while there is still some noise in the samples.}
      \label{tbl:experiments_mix}
\end{table}

\subsection{The Difference between Generative and Traditional Data Augmentation}

As generated dynamic images can be viewed as a kind of data augmentation, we compare it to conventional data augmentation methods of videos, corner cropping with scalejittering (CCS) and horizontal flipping (HF). The results are shown in Table \ref{tbl:experiments_augmentation}. We can see that the generated images by GAN provides a significantly improvement in the performance, because the generated samples could smooth the decision surface and play a role of regularization. 

\begin{table} 

    \centering
    \setlength{\tabcolsep}{6.5mm}{
    \begin{tabular}{llllll} 
    \toprule
    CCS & HF & GAN & Accuracy \\
      \midrule
      \checkmark &  &   & 70.28\% \\
       & \checkmark &   & 69.69\% \\
       &  & \checkmark  & 71.01\% \\
       & \checkmark & \checkmark  & 72.85\% \\
      \checkmark &  & \checkmark  & 76.60\% \\
      \checkmark & \checkmark &   & 76.38\% \\
      \checkmark & \checkmark  & \checkmark  & \textbf{79.57\%} \\
      \bottomrule
      \end{tabular}
      }
      \caption{Comparison of different data augmentation techniques with proposed method.}
      \label{tbl:experiments_augmentation}
\end{table}

\subsection{Sample Insufficiency Scenario}

It is very common in industrial applications to classify some niche videos, which are not many types with limited samples. Thus we evaluate the proposed method on HMDB4 and UCF4. There are three baselines: a) the model trained with dynamic images obtained by sparse sampling (SSDI). b) the model trained with dynamic images obtained by sparse sampling and randomly selected generated dynamic images (SSDI+RND-GAN). c) Using traditional self-paced learning method to replace random selection in b) (SSDI+TSP-GAN). The proposed method is denoted as SSDI+SPS-GAN.

\begin{figure}[ht]

    \centering
    \includegraphics[width=0.43\textwidth]{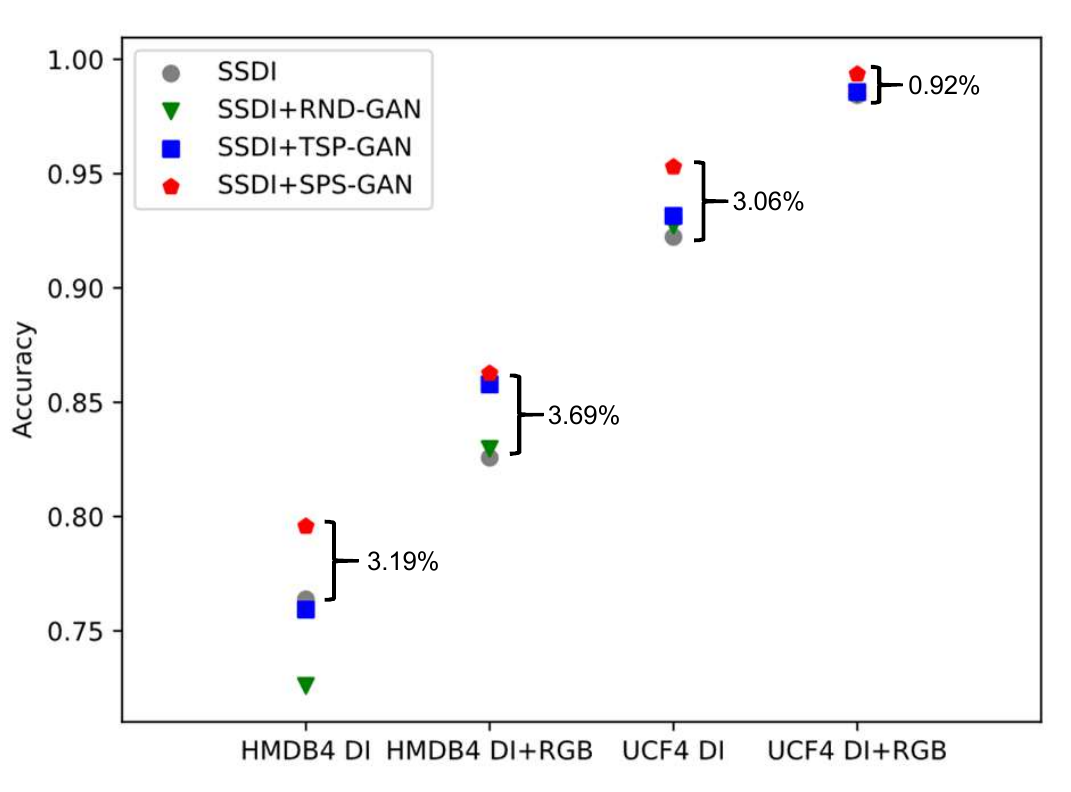}
    \caption{We conduct several experiments to validate the proposed method. The red markers represent the results of proposed method and the other are results of three baselines. DI is the abbreviation of dynamic image and the RGB is the color image.}
    \label{fig:hmdbucf4results}

\end{figure}

The results are shown in Figure \ref{fig:hmdbucf4results}. We can see that if we randomly select several generated samples for training, the accuracy may be impacted due to severe noise in some samples, where the accuracy is lower than the SSDI when we randomly select generated images for training in HMDB4 DI. As for self-paced learning, when we use the loss value to select the generated samples of HMDB4 DI, which contains relatively more noisy images than UCF4 DI, the performance is slightly lower than SSDI. We think that this phenomenon is caused by the gap between the generated images and the real images, and this difference is not taken into account during the training process, resulting in unstable losses of the classifier when classifying the generated images. After adding the SPS, we observe improvement of $+3.19\%$ (from $76.38\%$ to $79.57\%$) on HMDB4 DI and $+3.06\%$ (from $92.23\%$ to $95.29\%$) on UCF4 DI, which proves the effectiveness of our algorithm. Since the accuracy of Temporal ConvNet is improved, the fusion of Spatial ConvNet and Temporal ConvNet gets enhanced in the end.

\begin{figure}[ht]

    \centering
    \includegraphics[width=0.43\textwidth]{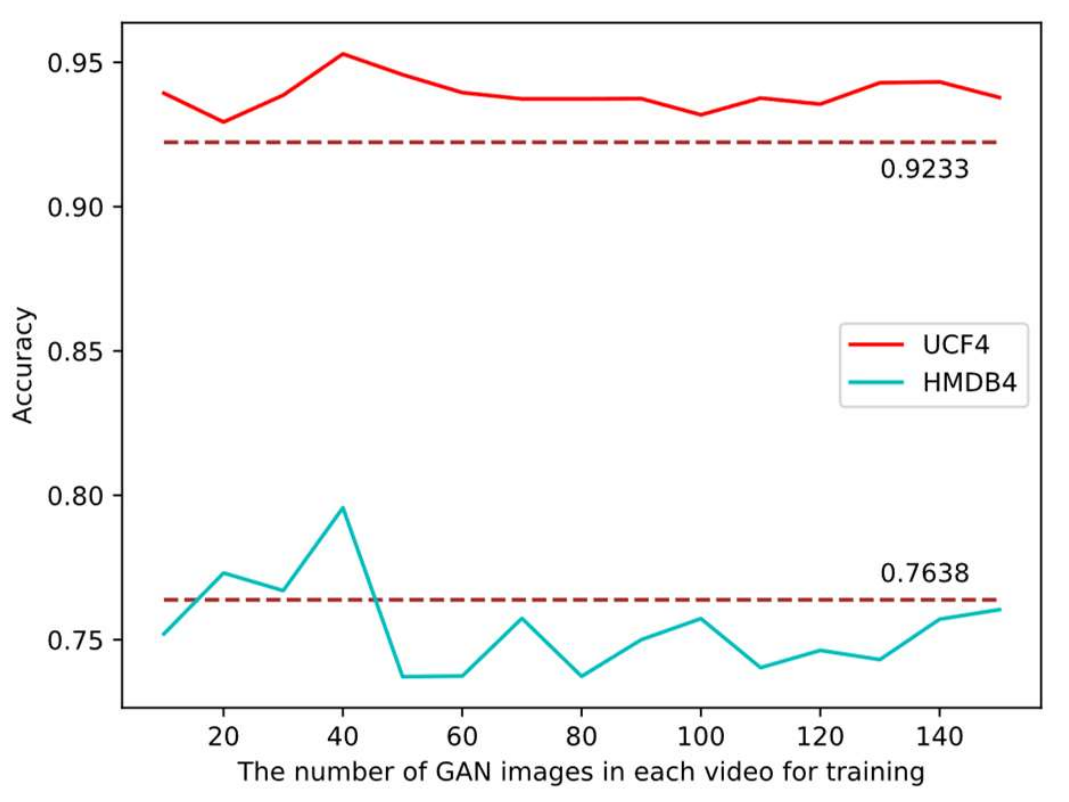}
    \caption{The horizontal axis is the number of GAN pictures used for training and 
    The vertical axis is the corresponding accuracy of temporal stream. The brown horizontal line represents the performance of the baseline which does not utilize the generated images.}
    \label{fig:ucf4numresults}

\end{figure}

\textbf{The impact of using different numbers of GAN images during training} We also evaluate how the number of generated dynamic images affects classification performance. From the Figure \ref{fig:ucf4numresults}, we find that the performance is always better when using the generated dynamic images in UCF4, while utilizing too much generated images in HMDB4 impacts the results. This is not surprising as the videos in HMDB4 are more complicated, thus the quality of generated dynamic images are poorer than that of UCF4. If we select too much images for the training process, there would be some generated images of poor quality to be added in the training set, which hamper the performance in the end. In addition, since there are lots of dynamic images obtained by sparse sampling in the training set, the performace will not drop too fast even with the noisy generated samples. 

\subsection{Sample Imbalance Scenario}

Sample imbalance is also a difficult problem in the video classification. In order to evaluate our method on the problem, we combine the HMDB4 and the other classes in HMDB51 as HMDB5$1^*$. 
UCF10$1^*$ is obtained by the same way with UCF4 and the other classes in UCF101. We train the network on HMDB5$1^*$ and UCF10$1^*$ and show that our method can achieve comparable results with methods trained on HMDB51 and UCF101. 

\textbf{The GAN images improve the baseline} We compare our method with the model trained with real dynamic images (SSDI). The results are shown in Table \ref{tbl:compare_UCF101}. The overall results are slightly enhanced as there are only four classes that are augmented. However, the accuracy of the four classes are imporved by a large range. The results of UCF$4^*$ and HMDB$4^*$ in Table \ref{tbl:compare_UCF101} is the accuracy of four classes in UCF4 and HMDB4 when the network is trained with UCF10$1^*$ and HMDB5$1^*$. The performance is increased by \textbf{7.2\%} of UCF$4^*$ and \textbf{6.8\%} of HMDB$4^*$ respectively. 

\begin{table}[htbp]
    \centering
    \setlength{\tabcolsep}{1mm}{
    \begin{tabular}{cccccc} %
    \toprule
      Method  & UCF10$1^*$ & UCF$4^*$ & HMDB5$1^*$ & HMDB$4^*$ \\ 
      \midrule
      SSDI  &  73.8\% & 59.4\% & 42.0\% & 7.08\%\\
      SS.+SP.  & 74.4\% & \textbf{66.6\%} & 42.5\% & \textbf{14.6\%}\\
      SSDI+RGB  & 90.9\% & - & 59.7\% & -\\
      SS.+SP.+RGB& \textbf{91.3\%} & - & \textbf{60.4\%} & -\\
      \bottomrule
      \end{tabular}}
     
      \caption{The results on UCF101 and HMDB51 in the case of sample imbalance. The SS.+SP. is the abbreviation of the proposed method SSDI+SPS-GAN. }
      \label{tbl:compare_UCF101}
\end{table}






\textbf{Comparisons with the state-of-the-art} Table \ref{tbl:compare_state_of_the_arts} shows the comparisons with state-of-the-art methods \cite{donahue2015long,srivastava2015unsupervised,Tran_2015_ICCV,zhu2016key,carreira2017quo} on both HMDB51 and UCF101 datasets. Note that our method trained on imbalanced HMDB5$1^*$ and UCF10$1^*$ is also comparable with some existing methods, which demonstrates the potential values for sample imbalance exploration when the sufficient training samples are difficult to obtain in certain type of videos. DI Net \cite{bilen2016dynamic} is the closest work with our approach, which also used the dynamic image to train the temporal stream network. As they do not use the generative data augmentation in the temporal stream, the accuracy is unsatisfactory. 

\begin{table}[ht]
    \centering
    \setlength{\tabcolsep}{1mm}{
    \begin{tabular}{cccccc} 
    \toprule
    Method & UCF101 & HMDB51  \\
      \midrule
      LSTM-composite  & 75.8\% & 44.1\% \\
      C3D ensemble & 85.2\% & - \\
      Two-Stream  & 88.0\%  & 59.4\%   \\
      Key Volume Mining  & 93.1\%  & 63.3\% \\
      TSN  & 94.2\% & 69.0\% \\
      Two-Stream I3D  & 98.0\% & 80.7\% \\
      \midrule
      DI Net  & 76.9\% & 42.8\% \\
      DI Net + IDT  & 85.1\% &  \textbf{65.2\%} \\
      Ours$^*$ & \textbf{91.3\%} & 60.4\% \\ 
      \bottomrule
      \end{tabular}}
      
      \caption{A comparison with the state-of-the-art. The method with * means that it uses the imbalanced training set of HMDB51 and UCF101. The results show that our method outperforms some existing methods with fewer training samples.}
      \label{tbl:compare_state_of_the_arts}
\end{table}

\section{Discussion}
\textbf{The applicability of our method in sample insufficiency scenario} 
The goal of our method is to perform data augmentation under the circumstance of insufficient samples. It can be applied to any network for video classification with good expansibility. We  validate our generative data augmentation method using the TSN network as it works well on the scenario of sample insufficiency. We compare the our method with some 3D-CNN methods-C3D, R3D and I3D, which achieved state-of-the-art results in the field of video classification. The results are shown in Table \ref{tbl:experiments_3D}. The accuracy of 3D CNN methods are far lower than TSN as these methods require huge number of samples due to large parameters needed to optimize. 


\begin{table}[ht] 

    \centering
    \setlength{\tabcolsep}{1mm}{
    \begin{tabular}{ccccccc} 
    \toprule

    Method & C3D     & R3D     &  I3D    & TSN & Ours  \\
    \midrule
    HMDB4  & 70.24\% & 64.08\% &  68.10\% & 82.57\% & \textbf{86.26\%} \\
    UCF4   & 84.32\% & 69.18\% &  90.98\% & 98.43\%  & \textbf{99.35\%} \\
    \bottomrule
    \end{tabular}
    }
    \caption{We compare our method with some 3D cnn methods. It can be seen that simple network works better when the training samples are insufficient, which explains why we choose the TSN network as the backbone.}
    \label{tbl:experiments_3D}
\end{table}

\textbf{Benefits of rank pooling in dynamic image for video augmentation} The major flaw of dynamic images is loss of information when compressing the videos. However, the compression by rank pooling maintains the important features of the video and removes the tiny details or the static background. After removing the interference factors, the GAN could better understand the video and generate samples with categorical information. In addition, our method adopts sparse sampling strategy to extract different features of the subvideos, which ensures that the important motion information could be obtained.

\section{Conclusion}

In this paper, we propose an effective data augmentation method for videos. We utilize the generated dynamic image to alleviate the lack of sufficient training samples as well as regularize the network. In order to reduce the effects of noise in the generated samples, we design the Self-Paced Selection method to automatically select the high-quality samples, adding them into the training set. Our method has good expansibility and can be incorporated into existing methods as a module. 


\bibliography{egbib}
\bibliographystyle{aaai}

\end{document}